\documentclass[letterpaper]{article} 
\usepackage{aaai2027}  
\usepackage{times}  
\usepackage{helvet}  
\usepackage{courier}  
\usepackage[hyphens]{url}  
\usepackage{graphicx} 
\usepackage{natbib}  
\usepackage{caption} 
\usepackage{booktabs}
\usepackage{placeins}

\title{Version- and Scope-Aware Question Answering over Normative Documents:\\
A Deployed System and an End-to-End Evaluation at Production Scale}

\author{
    Liuyin Wang\textsuperscript{\rm 1},
    Shuaipeng Jin\textsuperscript{\rm 1},
    Jiwei Shi\textsuperscript{\rm 1},
    Jensen Hsu\textsuperscript{\rm 1,2}\corresponding
}
\affiliations{
    \textsuperscript{\rm 1}Beijing Caizhi Technology Co., Ltd., Beijing, China\\
    \textsuperscript{\rm 2}dknownAI, Beijing, China\\
    \{wangliuyin, jinshuaipeng, shijiwei, xjj\}@czkj1010.com
}

\begin{document}
\maketitle

\begin{abstract}
Correctly answering a question grounded in normative documents often depends on information outside any single passage: whether the retrieved document is the version currently in force; whether it applies to the jurisdiction, subject (such as an institution or applicant), and date at issue; and whether each normative claim can be traced to its supporting source text. Hosted retrieval services have substantially lowered the engineering cost of building an initial system over such corpora, making ``upload the documents and ask'' a common default. We evaluate this default on approximately 73{,}000 candidate normative documents supplied to a production deployment. The evaluation uses a stratified sample of 200 questions from our published benchmark, with a gold source document for every question; the released sampling rule reads no system outputs or scores. We compare the hosted service with a governed system that resolves version and scope through explicit rules before generation. The governed system scored 97.7 overall, while the hosted service scored 88.1, a gap of 9.6 points computed from unrounded means. The question set, the answer text evaluated for both systems, the scores, and the scripts used to reproduce the reported benchmark statistics are public. The governed configuration has operated as a commercial product since January 2026 and serves 1{,}126 registered users; named customer organizations include Zhipu AI and Lecheng Health. By mid-April 2026, it had reached roughly 100{,}000 calls per workday.
\end{abstract}

{\makeatletter\def\do@url@hyp{}\makeatother
\begin{links}
    \link{Deployed system}{https://yun.dknowc.cn/wlcb/dknowc-chat/}
    \link{Benchmark and code}{https://github.com/ASI2030/normative-qa-benchmark/tree/main/subset-200}
\end{links}}

\section{Introduction}\label{sec:intro}

A staff member at a Chinese public-service window is asked a question with a
definite answer: whether a firm qualifies for a talent subsidy, what the filing
deadline is, and which authority has jurisdiction. The answer is in a particular
article of a particular normative document among tens of thousands issued by
national, provincial, and municipal bodies. Answering requires finding the
document that actually applies, confirming that it is the currently effective
version rather than an older version it replaced or a newer version that has
replaced it, and confirming that its scope covers this applicant, this region,
and this date. An error here is not merely a poor search result; it is an
incorrect normative statement made with the authority of the service window. It
can then propagate---repeated by the staff member, acted on by the applicant,
and perhaps corrected only after a deadline has passed.

Much of the information that separates a correct citation from a hazardous one
is not contained in the text of a single passage. Two documents may be almost
identical in wording but differ exactly on the point that decides the current
case: one has been repealed, one applies to a neighboring city, or one covers a
different class of applicant. Effective status, issuing authority, applicable
region and period, and amendment and repeal relations are metadata or
cross-document relations, and passage-level semantic similarity provides no
mechanism for identifying them \citep{lewis2020rag, karpukhin2020dpr}. Chinese
normative documents further weaken surface textual signals: among the 191
distinct source documents involved in the question set, many share generic
title frames such as ``Notice on \ldots,'' often differing only in date or
issuing authority.

At the same time, hosted retrieval services have substantially reduced the
engineering cost of building an initial system. A team can upload a corpus to a
service such as Gemini File Search \citep{googlefilesearch} and receive grounded
answers with citations without implementing chunking, embedding, indexing, or
reranking. A natural default is therefore to upload the documents and ask. What
this default delivers is often demonstrated on a showcase corpus rather than on
the operator's production-scale candidate document set.

We evaluate on a production-scale document set containing near-duplicate
documents and superseded versions. Using 200 questions, each with a gold source
document, we ran two systems over approximately 73,000 candidate normative
documents---including documents issued by bodies at different levels and the
higher-level normative documents on which they rely, namely the candidate set
supplied to the deployment---and evaluated them under the same scoring protocol
\citep{zheng2023judge}.
One system is the hosted service; the other is DeepKnown, which resolves version
and scope by explicit rules before generation. DeepKnown scored 97.7 overall,
and the hosted service scored 88.1, a difference of 9.6 points computed from unrounded means.

This is a comparison between two deliverable systems, as it would be faced by
an operator choosing between them: the systems differ not only in whether they
resolve version and scope explicitly, but also in retriever, generator, and
index. Questions and expected answer points were constructed through a
rule-guided, AI-assisted process, grounded in original evidence, and reviewed
by the business owner (see the Evaluation Methodology section).

The governed configuration has run in production since January 1, 2026 and, as
of September 7, 2026, served 1,126 registered users; named customer organizations include Zhipu AI and
Lecheng Health. Workday call volume grew from roughly 10,000--20,000 calls per
day in January to roughly 100,000 by mid-April 2026. The operator also reports
lower token use during the knowledge-interpretation stage; because the comparison
baseline and measurement rule are not reported, we treat this observation as
qualitative rather than as a quantified comparative result. These operational
records document live use, user scale, and call volume;
answer quality is measured separately on the benchmark, under controlled
conditions in which both systems face the same questions and the same candidate
corpus (see the Deployed Application section).

Our contributions are: (i) a deployed normative-document question-answering
system in which version, scope, and effective status are resolved by an
explicit governance layer rather than treated as properties expected to emerge
from retrieval; (ii) a benchmark of 200 Chinese normative-document
questions---covering simple, complex, and partial-answer questions---released
with expected answer points, gold source documents, the answer text supplied to
the judge, citation counts, scores, and judge rationales for both systems, plus
the sampling rule (\texttt{scripts/make\_subset.py}) and a script
(\texttt{scripts/reproduce\_200.py}) that recomputes the benchmark statistics
reported in this paper; and (iii) an end-to-end evaluation at the scale of the
candidate document set supplied to the deployment, decomposed by question type
and credit outcome.

\section{The Deployed Application}
\label{sec:deployment}

The application answers questions about \emph{normative documents}. Within the
empirical scope of this paper, these documents mainly include notices, service
catalogs, implementing measures, product terms, and fee schedules issued,
revised, withdrawn, or repealed by enterprises and government bodies. The application
neither formulates these norms nor decides individual cases on behalf of the
responsible entity; its job is to put the currently effective, scope-matched
clause in front of the person who must answer or act on it. The deployment and
evaluation described in this paper use a candidate set of
normative documents drawn from government public-service operations. Two roles use the service: service-window and hotline staff,
who must answer within the span of a single conversation, and normative-document
consultants and service caseworkers, who compose written replies or implement
services under the applicable norms and need the citation as much as the answer
itself. Both groups are answerable for what they say, while the materials they
cite were not written by them. Unaided, the task is manual retrieval over a
corpus that no one can master from memory: the 200 benchmark questions alone are
associated with 191 distinct source documents.

\textbf{Deployment record.}
The configuration described here went live on 1 January 2026 as an upgrade to a
service already in operation. As of 7 September 2026, it had 1{,}126 registered
users; named customer organizations include Zhipu AI (z.ai), a commercial large-language-model service
provider, and Lecheng Health, a health-services organization. The service is
offered as a commercial product and runs at
\url{https://yun.dknowc.cn/wlcb/dknowc-chat/}; registration is self-service.
Reviewers unable to complete registration may request access from the
corresponding author. Service-platform monitoring covers 1 January to 19 April
2026 and distinguishes workdays, weekends, and public holidays. Workday volume
rose from roughly 10{,}000--20{,}000 calls per day in January to roughly
100{,}000 by mid-April, a five- to tenfold increase; a workday near the end of
the observation window carried about 98{,}000 calls, which is the level reached
by April rather than an average since launch. Two properties of the volume
record are important. First, volume falls during the Spring Festival holiday in
mid-February and on every weekend, which is consistent with load following the
work calendar rather than a batch or manually constructed schedule. Second, volume
rises gradually over fifteen weeks, showing a sustained accumulation of use. A
call is a service invocation, not necessarily a distinct end-user question, so
we report call volume rather than per-user rates.

\textbf{Operational evidence and benchmark evidence.}
Operational performance data help establish that the system is a deployed
application rather than only a prototype. The available records document live
operation during the monitored period, user scale, and call volume, together
with one operator-reported operational benefit: lower token use in the
knowledge-interpretation stage. Because the comparison baseline and measurement
rule are not reported, we treat this as qualitative operational evidence rather
than a quantified comparative result. Answer quality is measured
separately on the benchmark, under the controlled conditions set out in the
Evaluation Methodology section, so that the two systems face the same
200 questions and the same input candidate document set. The
improvement we claim is the benchmark result of 97.7 versus 88.1 overall in the
production-scale setting of the candidate normative-document corpus, which can
be recomputed from the public per-question records.

\textbf{A single interaction.}
Questions often arrive underspecified: a year, a jurisdiction, or an applicant
category may be missing, because the asker does not know that these elements
determine the answer. Retrieval runs over segmented documents, but the retrieval
results themselves do not determine whether a document may be used as the basis
for an answer. A governance layer then filters candidates along two dimensions:
\emph{version-awareness}, whether a clause belongs to the currently effective
version of a document rather than a superseded, expired, or repealed version;
and \emph{scope-awareness}, whether the document applies to the jurisdiction,
applicant category, and time period asked about. Generation is confined to the
admissible evidence set, and each assertion carries an inline marker pointing to
the clause that supports it. When this evidence is insufficient to determine an
answer, the system abstains and states the missing information rather than
supplying the nearest merely plausible figure \citep{feng2024abstain}.

\textbf{Alternatives in the design space.}
At the stages just described, three lower-cost designs were available; our
objection to each is architectural, and the AI Approach and Design Rationale
section develops the argument. Version and scope could be left to
general-purpose retrieval \citep{lewis2020rag}, which reads them as ordinary
words in the passage---this alternative is the hosted baseline we measure.
Lineage could be written as a provenance header in the chunk text rather than as
typed fields; this would require no schema and would work in any hosted index.
Answerability could be left for the generator to declare, rather than being
gated by an evidence test over the filtered set
\citep{wen2024knowyourlimits}. Each design is cheaper to build, but its cost
appears later in the pipeline.

\textbf{Why the error budget is small.}
A figure stated at a service window is taken as an official commitment: a
citizen files materials, pays, or waits on that basis, and reversing the
consequence requires an administrative proceeding, not an edit to text. In this
setting, the dangerous failure of a fluent system is not silence but a
well-formed wrong answer wearing a seemingly real citation---a failure already
documented for general-purpose models on legal questions
\citep{dahl2024legal} and, in an audit of commercial legal research tools, for
systems that provide citations while answering
\citep{magesh2025hallucinationfree}. Answers from the system with the
governance layer therefore begin with a fixed notice directing the reader to
consult the cited original text, and each assertion carries an inline marker
pointing to the clause that supports it.

\section{Normative Knowledge as an Application Requirement}
\label{sec:normative-knowledge}
\suppressfloats[t]

Understanding the requirements of this application first requires distinguishing naturally occurring general knowledge from human-designed normative knowledge. General knowledge includes natural regularities, scientific facts, linguistic competence, and relatively stable common knowledge; its correctness can usually be supported by objective validation, empirical consensus, or logical reasoning. Normative knowledge, by contrast, is a human-created knowledge system that is designed, stipulated, codified, or organized within a specific domain: a set of formal documents provides the authoritative basis for a correct and comprehensive understanding of that domain, and any execution, judgment, or answer must trace back to the source text of documents whose scope actually applies. It is not limited to laws or public policy. Industry standards, service catalogs, operating procedures, product terms, customer-service rules, product manuals, and standard solutions can all be normative documents in this sense.

The correctness of normative knowledge depends not only on textual semantics but also on the issuing authority, effective status, effective date, applicable jurisdiction and subject, and replacement and hierarchical relationships among documents. Identical or nearly identical wording can lead to different conclusions under different versions, authorities, and scopes. Providing normative-knowledge services is therefore not a matter of extracting an answer that is merely ``probably correct'' from documents. It requires determining which documents and provisions constitute usable evidence for the current request while preserving the textual basis and accountability chain.

\begin{figure*}[t]
    \centering
    \includegraphics[width=\textwidth]{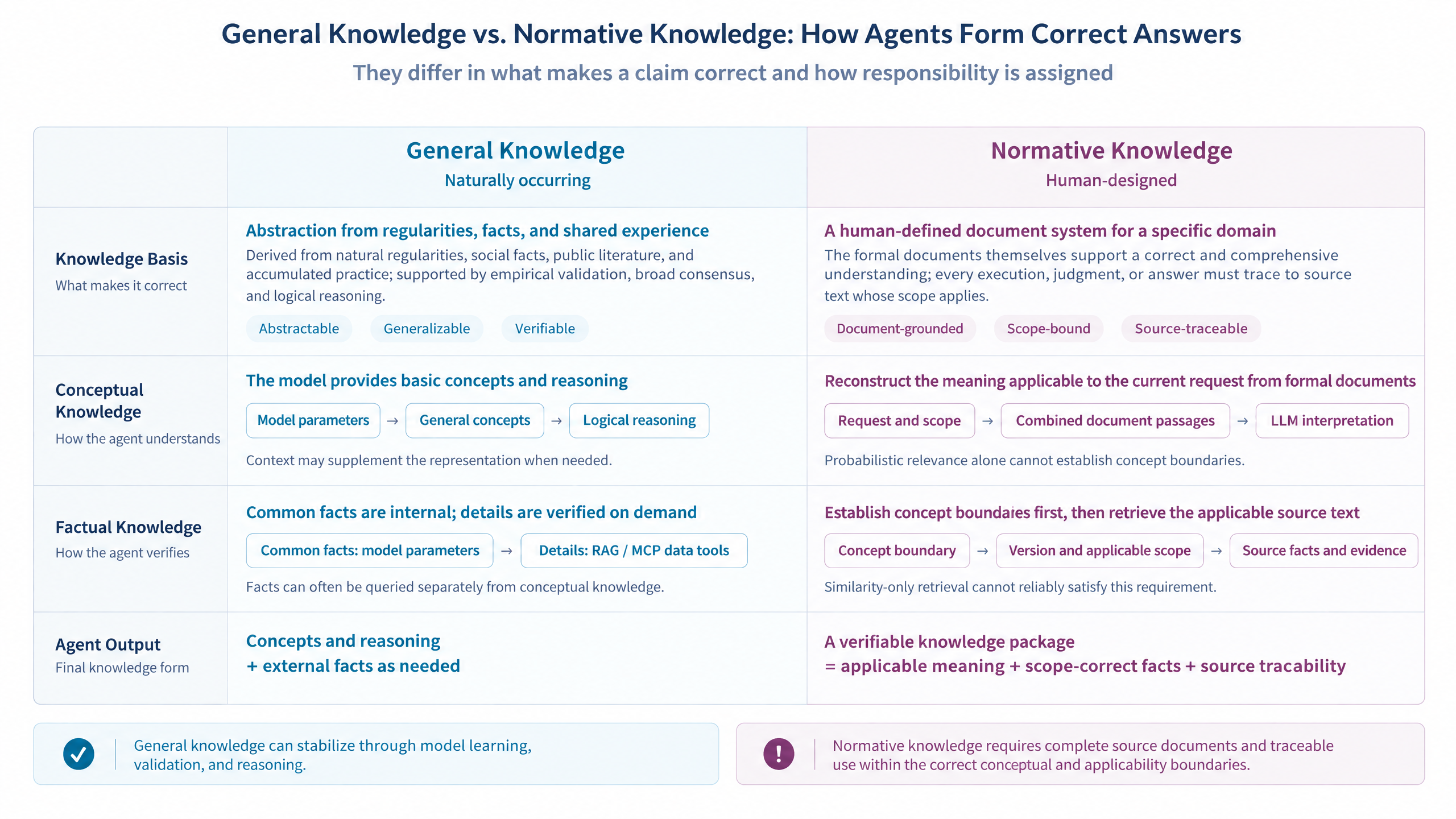}
    \caption{General and normative knowledge: their defining properties and the paths by which an agent can form a correct understanding.}
    \label{fig:knowledge-types}
\end{figure*}

This distinction determines how an agent can form a correct understanding. General knowledge can rely primarily on concepts and reasoning abilities formed by the model, with detailed facts supplemented as needed by RAG or data tools. Normative knowledge, by contrast, requires organizing document passages whose scope matches the current request and retrieving facts within accurate conceptual boundaries, effective versions, and applicable scopes. Even when similarity-only retrieval finds textually close passages, it cannot reliably establish that those passages are currently in force and applicable to the question at hand.

A system for normative knowledge therefore faces two core requirements. First, the system must preserve complete normative documents, their effective versions, and their applicability boundaries, so that it can handle the continuing addition, amendment, replacement, and repeal of documents; detached summaries or internal states are not sufficient. Second, during retrieval and use, any content that lacks source-text support, is no longer in force, or falls outside the applicable scope must be excluded from the conclusion. Every normative claim should be traceable to its issuing authority, source document, specific provision, version, and applicable scope, supporting verification and audit. These requirements directly motivate the system design in this paper. The next section explains how DeepKnown implements them as executable mechanisms through ingestion-time governance, hard predicates, version lineage, and an external evidence test.

\FloatBarrier

\section{AI Approach and Design Rationale}
\label{sec:approach}

In the broad sense, both systems are retrieval-augmented \citep{lewis2020rag};
they differ in where, and in what form, the constraints that determine whether a
document applies at all---jurisdiction, issuer, level of issuing authority, effective date,
supersession, and regulated subject---are represented and enforced. The hosted service does not expose application-specific version-lineage,
scope, or effective-status logic as enforceable predicates \citep{googlefilesearch}.
In the evaluated configuration, these properties were not represented by an
additional operator-defined governance layer, and any filtering performed
inside the managed ranker was not inspectable by the operator. The governed system extracts
these constraints at ingest, stores them as typed fields alongside the text,
and, when extraction confidence is sufficient, enforces them as predicates
\emph{before} ranking.

\paragraph{Governance occurs at ingestion.}
Each file passes through attribute extraction (issuer, document number,
promulgation and effective dates, and document type), version-lineage
resolution, and scope annotation along three axes: geography, time, and
regulated subject. Extraction is model-assisted and human-reviewed. Segmentation
follows the document's own article structure rather than a fixed token window,
so a chunk inherits a coherent set of attributes instead of straddling two
articles with different effective dates.

\paragraph{Decision 1: hard predicates rather than similarity signals.}
The less costly design would use a single dense index and treat metadata as a
soft weight at ranking time. We rejected that design because, in a candidate
normative-document corpus composed of documents issued by authorities at
different levels, many pairs are nearly identical in wording but disjoint in
scope---successive versions of the same measure, or parallel measures from
different municipalities. For such documents, no fixed similarity margin can
separate the document that actually governs the question from one that merely
reads alike; among the 191 source documents behind the question set, many share
generic title frames such as ``Notice on \ldots''. A soft weight can be offset
by other ranking signals; a hard predicate cannot. The cost we accepted is
possible loss of recall: an error in attribute extraction can remove the
correct document, turning a ranking error into a missing-evidence error. We
relax filters one dimension at a time in a fixed order. If the relaxation
options are exhausted and admissible evidence still cannot be assembled, the system
does not output an unsupported conclusion.

\paragraph{Decision 2: version lineage as a field, not as chunk text.}
Adding a provenance line to the chunk text (``issued in 2019; superseded by
X'') requires no schema and works in any hosted index. We rejected it for three
reasons. First, it requires the generator to trust that line rather than a
lexically stronger match elsewhere in the context; prior reports indicate that
this behavior degrades in the presence of distractors and long contexts
\citep{cuconasu2024noise,liu2024lost}. Second, supersession is discovered
\emph{after} the superseded document has been ingested, so the provenance-header
approach would require re-vectorizing an entire lineage whenever a new file
arrives, whereas the field-based approach requires updating only one relation
edge. Third, free text expresses partial supersession poorly. Question G-044 in
the released set has this form: a 2023 municipal guideline expired in May 2025,
after which a separate 2024 document governed. Both systems answered this
question correctly, so it illustrates the task form rather than a result. The
price is a schema to maintain and a fallible extraction step; accordingly,
attributes are used to exclude documents only when confidence exceeds a
threshold, and otherwise are used only to adjust ranking.

\paragraph{Decision 3: an external evidence test rather than model
self-assessment of evidence sufficiency.}
The common alternative leaves the decision about whether a conclusion has been
established to the generator---prompting it to check its own support, or
training it to emit self-reflection tokens \citep{asai2024selfrag}. We instead
define evidence sufficiency as a property of the filtered candidate set: the
system enters answer generation only when the retained evidence satisfies the
question's scope and version constraints. A generator's internal self-judgment
is itself affected by retrieval noise, and models do not reliably report the
boundaries of their own competence \citep{wen2024knowyourlimits,kirichenko2025abstentionbench}.
In this 200-question evaluation, in which every question had a gold source
document, the hosted service returned no output for 6 questions and returned answers
without any citation for 2 questions; all were assigned zero under the
prespecified rules. These records illustrate that producing text, and producing fluent text,
cannot by itself substitute for an independent test of whether the external
evidence is sufficient and applicable.

\paragraph{What the design costs.}
The governance layer costs per-domain schema definition, ingest latency, and a
review loop that does not disappear---costs that some applications would
reasonably decline in exchange for the operational simplicity of a hosted
index. On this corpus, the governed system scored 97.7 overall and the hosted
service scored 88.1, a gap of 9.6 points. The governed system nevertheless did
not receive full credit on every item: it received full credit on 189 of the
200 questions, while its remaining 11 items---3 simple, 6 complex, and 2
partial-answer questions---received partial credit. None received zero. The governed configuration checks evidentiary scope and
version applicability before generation. Our claim is not that governance is
always warranted, but that in this domain, at the scale of the candidate corpus supplied to the
system, the constraints that determine whether a document
applies are worth enforcing where they can be inspected, rather than leaving
them to an opaque ranker.

\section{Evaluation Methodology}\label{sec:method}

\subsection{Systems, Corpus, and Controls}\label{sec:systems-controls}

We compare two systems using the same candidate set of normative documents. DeepKnown interposes an explicit governance layer between retrieval and generation (see the AI Approach section). Google Gemini File Search (\texttt{gemini-3.5-flash} with \texttt{gemini-embedding-2}) is a hosted RAG service in which ingestion, retrieval, and grounding are managed by the provider. In the configuration evaluated here, we used the managed retrieval pipeline without adding application-specific version, scope, or lineage rules. The comparison is therefore between two deliverable systems in the form in which an operator would choose between them: DeepKnown and the hosted service differ in governance layer, base model, ingestion pipeline, index, and answer template.

The evaluation has a single experimental condition. Both systems received the same set of approximately 73,000 candidate normative documents and were given the same 200 questions. After ingesting the files, each system applied its own automated pipeline for parsing, quality assessment, and knowledge representation; a small number of files that were of very low quality, unstable to parse, or otherwise inadmissible may have been automatically excluded. This figure denotes the size of the shared candidate document set, not a claim that every file entered both systems' final effective indexes. Corpus size is not an experimental variable here but a fixed operating condition; all reported benchmark results describe system behavior under that condition. The released records confirm that the two systems received identical question text, type labels, expected answer points, and gold source documents.

\subsection{Question Set}\label{sec:questions}

The benchmark contains 200 questions: 106 \emph{simple} questions, answerable from a single provision; 81 \emph{complex} questions, spanning multiple constraints, conditions, or documents; and 13 \emph{partial-answer} questions, for which the corpus supports part of the answer and a correct response must explicitly distinguish supported from unsupported parts. Every question has a gold source document and expected answer points.

These 200 items are a stratified random sample of the 420 answerable items in our published 460-item benchmark release. The strata are question type and the jurisdiction named in the gold source document, both fixed before either system was run; allocation across strata is proportional, and the random seed was fixed before either system's outputs or scores were inspected. The selection reads no scores, answers, citations, or judge rationales, so which items it retains cannot depend on how either system performed on them. The selection rule is released as \texttt{scripts/make\_subset.py} in the benchmark repository; run with \texttt{--check}, it regenerates \texttt{subset-200/} byte for byte from the published release.

Questions and expected answer points were constructed through a rule-guided, AI-assisted process, grounded in original evidence, and reviewed by the business owner. The business owner defined the type criteria, the conventions for scope and current effectiveness, and the evidentiary boundary of an acceptable answer. The question-construction pipeline then used model assistance to build questions from extracted source passages, and the items were frozen by script. The full process was subjected, in sequence, to programmatic checks for source existence, scope, current effectiveness, title conflicts, evidence, and format, followed by human spot-checking and anomaly review.

\subsection{Scoring and Rule-Assigned Records}\label{sec:scoring}

Scoring has two layers. A deterministic rule layer---executed by the scorecard rather than by the judge---directly assigns scores to records that satisfy predefined objective conditions; all remaining records are scored by an LLM judge against the expected answer points with partial credit \citep{zheng2023judge}. The scorecard specifies that an answer that is empty or carries no citation receives $0.0$, because in knowledge-base question answering an answer without provenance counts as wrong. Among the 400 records (two systems $\times$ 200 questions), 8 records were assigned directly by this rule, all from the hosted service: 6 empty returns and 2 answers without any citation. No score is missing.

Reported figures are per-question means on a 0--100 scale. Scoring is not binary: 30 of the 400 judgments are fractional. Because both systems were evaluated on the same 200 questions, we quantify uncertainty for each difference using 95\% percentile bootstrap confidence intervals with resampling by question. The accompanying script recomputes every cell in Tables~\ref{tab:main}--\ref{tab:credit} from the public records and exits with a nonzero status on disagreement.

The judge is GPT-5.5, called through an OpenAI-compatible endpoint at temperature 0, and is not given an explicit system label. The prompt contains only the question, the first 1,500 characters of the answer, and the expected answer points; it excludes the system-identity field and the separate citation fields. The released records preserve exactly the answer text supplied to the judge. The provenance rule is therefore enforced by the scorecard, not by the model. The judge determines, item by item, whether each expected point is covered, and the score is the covered proportion. The same model, temperature, scoring formula, and rules were used for both systems; a prompt wording revision between batches was applied symmetrically to both systems, while the formula and rules remained unchanged, as recorded in the released protocol. Rule-assigned records can be identified by their rationale strings in the released data, the judge prompts are published with the protocol, and any released answer can be re-judged under another configuration.

\subsection{Why End-to-End Scores}\label{sec:end-to-end}

The primary metric is the answer the user actually receives, not a retrieval proxy such as gold-file hit rate. The two systems expose citations in different forms: the hosted service reports the number of grounded citations, whereas DeepKnown reports the title of the cited document. The released records preserve the evaluated answer text and the citation count used by the scorecard. For the failure modes of interest---superseded versions, scope mismatches, incomplete answer points, or inconsistency between citations and conclusions---hitting a relevant file is not sufficient to ensure that the final answer is correct; retrieval-side metrics \citep{es2024ragas} also cannot replace scoring the answer actually delivered to the user. We release the gold source document for every question to support answer-level verification; computing retrieval metrics would require additional retrieval traces that are not exposed in a common format by both systems.

\subsection{Reproducibility}\label{sec:reproducibility}

The 200-question set (\texttt{subset-200/}), the per-question records for both systems (the answer text supplied to the judge, citation counts, scores, and the rationale for each score), the rule that drew the set (\texttt{scripts/make\_subset.py}), and the script that recomputes the reported benchmark tables, confidence intervals, credit counts, scoring-layer counts, and distinct-source count (\texttt{scripts/reproduce\_200.py}) are all public in the benchmark repository, alongside the published 460-item release. Readers can rescore the released answer text, replace the judge, or recompute Tables~\ref{tab:main}--\ref{tab:credit} and the other benchmark aggregates checked by the script. The benchmark results complement the deployment record in the Deployed Application section.

\section{Results}
\label{sec:results}

Table~\ref{tab:main} reports per-question judge-score means on a 0--100 scale
over the candidate normative-document corpus: two systems, the same 200
questions. \texttt{scripts/reproduce\_200.py} in the benchmark repository regenerates every
cell in Tables~\ref{tab:main}--\ref{tab:credit} from the released per-question
records (\texttt{subset-200/results.json}); every other benchmark figure below comes from those same records. Scoring is
not binary---30 of the 400 judgments in this condition are fractional---so a
mean of 97.7 is a mean of per-question credit rather than a share of fully
correct answers \citep{zheng2023judge}. Intervals are 95\% percentile
bootstrap confidence intervals that resample questions, pairing the two
systems on each item.

\begin{table}[t]
\centering
\small
\setlength{\tabcolsep}{3pt}
\begin{tabular}{lrrrl}
\toprule
Question set & $n$ & DeepKnown & Gemini & Gap [95\% CI] \\
\midrule
\textbf{All questions} & \textbf{200} & \textbf{97.7} & \textbf{88.1} & \textbf{$+9.6$ [5.7, 13.8]} \\
\bottomrule
\end{tabular}
\caption{Mean judge score (0--100) on 200 questions over the candidate
normative-document corpus. Gap is DeepKnown minus Gemini, computed from
unrounded means.}
\label{tab:main}
\end{table}

\paragraph{The measured difference.}
DeepKnown scores 97.7 overall, compared with 88.1 for the hosted service, a
difference of 9.6 points; the paired bootstrap 95\% confidence interval is
[5.7, 13.8], which excludes zero. Because DeepKnown is already close to the top of the scale, the maximum
additional credit available to it is small: among the 200 questions, DeepKnown
receives full credit on 189 and zero credit on none, while the corresponding
counts for the hosted service are 166 and 15.

\begin{table}[t]
\centering
\small
\setlength{\tabcolsep}{3.5pt}
\begin{tabular}{lrrr}
\toprule
Type & $n$ & DK & Gem. \\
\midrule
Simple    & 106 & 99.0 & 91.1 \\
Complex   & 81  & 96.8 & 86.6 \\
Partial   & 13  & 92.3 & 73.1 \\
\textbf{All} & \textbf{200} & \textbf{97.7} & \textbf{88.1} \\
\bottomrule
\end{tabular}
\caption{Scores by question type on the candidate normative-document corpus.
DK is DeepKnown; Gem. is the hosted service; recomputed from the released
records.}
\label{tab:bytype}
\end{table}

\begin{table}[t]
\centering
\small
\setlength{\tabcolsep}{3.5pt}
\begin{tabular}{lrrrrrrr}
\toprule
 & & \multicolumn{3}{c}{DeepKnown} & \multicolumn{3}{c}{Gemini} \\
\cmidrule(lr){3-5}\cmidrule(lr){6-8}
Type & $n$ & full & part. & zero & full & part. & zero \\
\midrule
Simple    & 106 & 103 & 3  & 0 & 95  & 3  & 8  \\
Complex   & 81  & 75  & 6  & 0 & 65  & 9  & 7  \\
Partial   & 13  & 11  & 2  & 0 & 6   & 7  & 0  \\
\textbf{All} & \textbf{200} & \textbf{189} & \textbf{11} & \textbf{0} & \textbf{166} & \textbf{19} & \textbf{15} \\
\bottomrule
\end{tabular}
\caption{Credit breakdown by question type: items scored at full credit,
partial credit, and zero. Recomputed from the released records.}
\label{tab:credit}
\end{table}

\paragraph{Anatomy of the gap.}
Table~\ref{tab:credit} breaks each type into full, partial, and zero credit.
DeepKnown receives zero credit on none of the 200 questions; its 11 non-full
items all receive partial credit. The hosted service's 15 zero-credit items
fall into three groups: 6 empty returns, where the pipeline produced no
answer; 2 answers with no citation, scored zero under the provenance rule; and
7 cited answers that the judge found substantively wrong; those answers cited
13 documents on average. Asked for a prefecture's 2017 affordable-housing
targets, the hosted service cited 35 documents and returned county-level
figures inconsistent with the city-wide targets the source provision sets.
Asked for the talent allowance that Shenzhen's Longhua District
provides to a full-time doctoral graduate from a university ranked among the world's top 500, it
answered RMB 80{,}000, whereas the source provision sets RMB 200{,}000. Asked
for the weights assigned to renovation of old urban residential communities and shantytown redevelopment
in the allocation of central-government affordable-housing subsidy funds, it
gave 80\% for both, whereas the rule sets 30\% for renovation of old urban residential communities and 10\% for shantytown redevelopment. This type of failure is especially risky in a service-desk setting: each
answer is fluent and cited, but the conclusion is wrong. On partial-answer items---where the correct output
states what the corpus supports and marks what it does not support---DeepKnown
receives full credit on 11 of 13, compared with 6 of 13 for the hosted
service.

\paragraph{Where the difference appears.}
The split by type in Table~\ref{tab:bytype} shows the hosted service furthest
behind on \emph{partial-answer} items (73.1 against 92.3), then on
\emph{complex} items (86.6 against 96.8), and closest on \emph{simple} ones
(91.1 against 99.0). Partial-answer items are the hardest class for both
systems. At these sample sizes the three intervals---[7.7, 30.8] for
partial-answer items, [3.5, 18.0] for complex, and [3.2, 13.3] for
simple---all exclude zero but overlap one another, so the direction holds for
every type while the data do not establish a reliable ordering of the gap
sizes.

Partial-answer items are where the two designs differ most sharply. One
plausible explanation is structural: the correct output states what the corpus
supports and marks what it does not, which requires a decision about the
admissible evidence set rather than only the best-matching passage. A system
without an explicit admissibility test may complete the answer from the
retrieved context, whereas the governance layer is designed to stop at the
boundary of the evidence. On simple single-provision items, the gap is smaller
than on partial-answer items. This interpretation remains a hypothesis; testing
it would require per-item retrieval traces, which we leave to future work.

\paragraph{Scope.}
This is one question set, one document domain, one corpus, and one comparison
pair, scored under the scoring protocol described in the Evaluation Methodology
section. The comparison is whole-system: DeepKnown and the hosted service
differ not only in the governance layer but also in base model, ingestion,
index, and answer template, which is the comparison an operator choosing
between the two actually faces. What the result supports is this: on the
candidate normative-document corpus, with 200 questions each having a gold
source document, and under the scoring protocol described in the Evaluation Methodology section, the governed configuration
is 9.6 points higher overall and scores higher on simple, complex, and
partial-answer questions. The benchmark result is complemented by the deployment
record: the governed configuration has run in production since 1 January 2026
for 1{,}126 registered users; by April 2026, workday volume had reached
roughly 100{,}000 calls; and the operator reports lower token use during the
knowledge-interpretation stage, a qualitative observation for which no comparison
baseline or measurement rule is reported.

\section{Lessons Learned}\label{sec:lessons}

\paragraph{Governance belongs at ingestion, not in the prompt.}
\emph{Simple} questions---single-provision lookups in which the governing
document must compete with many surface-similar documents---still show a
7.9-point gap computed from unrounded means, while complex questions show a
10.3-point gap. The governed configuration resolves version and scope
constraints as typed predicates before ranking, and its simple-question score
is 99.0. The implication for other operators is that even apparently simple
lookups should not be treated as solved by similarity retrieval alone when
version and scope determine applicability. Schema definition is front-loaded
and domain-specific, but the resulting constraints are enforced at ingest
rather than left to the ranker.

\paragraph{Measure end to end under operating conditions.}
A buy-or-build decision should be made on the corpus the system will serve,
not on a demonstration set, and on the answer a user receives, not on retrieval
proxy metrics. Running both systems on the same candidate normative-document
corpus under the same scoring protocol made the value of the governed
configuration legible even to non-specialists: the overall gap on 200 questions
was 9.6 points, and the differences in the simple, complex, and partial-answer
categories all pointed in the same direction.

\paragraph{Distinguish empty returns, uncited answers, and cited errors.}
Among the 200 questions, the hosted service's 15 zero-point responses consisted
of 6 empty returns, 2 answers without citations, and 7 cited but
substantively incorrect answers. These three outcomes all receive zero in the
aggregate score, but they correspond to different system failure points. A
deployment evaluation should record output status, citation status, and answer
content separately, and should preserve the candidate set on which each
generated answer was based, so that errors can be traced to retrieval, version
or scope resolution, or generation.

\paragraph{Question sets need explicit business ownership and programmatic checks.}
Expected answer points for normative documents drift in ways that general
annotators will not detect: a point drawn from a superseded version, a point
outside the question's scope, or a character misrecognized during digitization.
Business-owner review establishes the conventions for question types, scope of
applicability, current effectiveness, and the evidentiary boundary of acceptable
answers; programmatic checks for source existence, scope of applicability,
current effectiveness, title conflicts, evidence, and format enforce those conventions
item by item. Both are necessary, and the latter is what keeps the question set
maintainable as the corpus changes.

\paragraph{Cost matters alongside accuracy.}
During five- to tenfold growth over the first fifteen weeks of deployment, the
operator reported lower token use in the knowledge-interpretation stage. Because
the comparison baseline and measurement rule are not reported, we treat this as
a qualitative cost observation; the benchmark also does not isolate its cause.
For the operator, however, the observation connects the quality argument with a
cost argument relevant to the budget owners responsible for service-window
operations.

\section{Related Work}\label{sec:related}

\paragraph{Question answering over legal and normative text.}
Legal and policy benchmarks emphasize reasoning and clause-level extraction:
LegalBench \citep{guha2023legalbench}, CUAD's annotated contract clauses
\citep{hendrycks2021cuad}, and, for Chinese, LawBench and LexEval, which
organize legal tasks by cognitive level
\citep{fei2024lawbench, li2024lexeval}. Administrative normative documents
stress a property that these suites do not isolate. An answer is acceptable
only if it is \emph{version-aware} and \emph{scope-aware}---keyed to the
document actually in force and to the jurisdiction and class of subject to which
the document applies. A working corpus of tens of thousands of such documents
accumulates superseded texts, near duplicates, and cross-jurisdictional texts
that match lexically while being normatively wrong. Closest in spirit is the
audit of commercial legal research tools by
\citet{magesh2025hallucinationfree}, which finds substantial residual
hallucination even when systems claim evidentiary support from retrieval
results. More broadly, fabricated legal citations in general-purpose models
have also been documented \citep{dahl2024legal}. Neither line of work reports
what a governed alternative achieves on the same questions.

\paragraph{Evidence support and end-to-end evaluation.}
Existing RAG evaluation includes combined assessments of retrieval relevance,
answer faithfulness, and generation quality \citep{es2024ragas}, as well as
metrics that check statement-level citation support \citep{gao2023alce}.
Other methods place the check during generation: self-reflection tokens inside
the generator \citep{asai2024selfrag}, or a lightweight evaluator that scores
retrieved passages and triggers correction \citep{yan2024crag}. The deployed
system studied here instead filters candidates before generation and directly
measures the final answer received by the user on a 200-question set in which
each question has a gold source document. Per-question records retain the evaluated answer text, citation count, score,
and judge rationale, supporting review of answer correctness and the provenance
rule. Retrieval-side analysis would require additional traces that the two
systems do not expose in a common format.

\paragraph{Hosted retrieval services.}
Hosted file-retrieval services---Gemini File Search \citep{googlefilesearch}
and OpenAI File Search \citep{openaifilesearch}---manage ingestion, chunking,
embedding, indexing, retrieval, and grounding behind an API or tool interface.
They are a reasonable default for building an initial system; public vendor
materials primarily present them as products rather than as systems
characterized through controlled, end-to-end comparative evaluation. The research evidence relevant to this comparison is mostly indirect. Reading studies show
that retrieval does not guarantee adequate evidentiary support: models
underuse evidence away from the context edges, and high-scoring but irrelevant
passages harm answers more than random passages do
\citep{liu2024lost, cuconasu2024noise}. The closest engineering account is a
qualitative failure taxonomy from three case studies
\citep{barnett2024sevenfailure}. These works do not report end-to-end answer quality for a hosted retrieval
service on an operator's own production-scale candidate document set, which is
the measurement needed for a buy-or-build decision.

\paragraph{Positioning.}
The retrieval components are standard \citep{lewis2020rag}; the innovation is
where the normative constraints are represented. Version, scope, and effective status are parsed as typed fields and, when
extraction confidence is sufficient, enforced as predicates before ranking
rather than left only as words in the chunk text. Our additional contribution is the setting in
which the measurement is taken: a commercial hosted service and a system in
production since January 2026, a fixed 200-question set, the candidate
normative-document corpus supplied to the deployed system, and scoring
of the answer that the user receives rather than a retrieval proxy. Under the scoring protocol described in the Evaluation Methodology section, the governed
system scores 97.7 overall, and the hosted service scores 88.1. The
comparison is whole-system: the two systems differ simultaneously in model,
index, prompting, and governance. That is precisely the comparison an operator
faces when choosing between a hosted index and a governed pipeline.

\bibliography{refs}

\end{document}